\documentclass[11pt,a4paper]{article}
\pdfoutput=1
\usepackage[hyperref]{naaclhlt2019}
\usepackage{times}
\usepackage{latexsym}

\usepackage{url}
\usepackage{rotating}
\usepackage{multirow}
\usepackage[inline]{enumitem}
\usepackage{subfigure}
\usepackage{amsmath}
\usepackage{paralist}
\usepackage{natbib}

\newlist{inlinelist}{enumerate*}{1}
\setlist*[inlinelist,1]{
 label=(\roman*),
}
\aclfinalcopy

\title{Expanding the Text Classification Toolbox with Cross-Lingual Embeddings}

\author{Meryem M'hamdi$^{\ddagger}$\thanks{This work has been conducted while being a student at Ecole Polytechnique F\'ed\'erale de Lausanne (EPFL) and master thesis intern in Swisscom (Switzerland) AG} , {\bf{Robert West$^{\dagger}$,  Andreea Hossmann$^{+}$, Michael Baeriswyl$^{+}$ and Claudiu Musat$^{+}$}}\\
$^{\ddagger}$ †
Information Sciences Institute \& Computer Science Department \\ University of Southern California (USC) \\ {\tt meryem@isi.edu} 
\\
$^{\dagger}$ Ecole Polytechnique F\'ed\'erale de Lausanne (EPFL) \\ {\tt \{firstName.lastName\}@epfl.ch}\\
$^{+}$Data, Analytics \& AI « --- » Swisscom AG \\
{\tt \{firstName.lastName\}@swisscom.com}  
}

\date{}

\begin{document}

\maketitle
\begin{abstract}
Most work in text classification and Natural Language Processing (NLP) focuses on English or a handful of other languages that have text corpora of hundreds of millions of words. This is creating a new version of the digital divide: the artificial intelligence (AI) divide. Transfer-based approaches, such as Cross-Lingual Text Classification (CLTC) - the task of categorizing texts written in \emph{different} languages into a common taxonomy, are a promising solution to the emerging AI divide. Recent work on CLTC has focused on demonstrating the benefits of using bilingual word embeddings as features, relegating the CLTC problem to a mere benchmark based on a simple averaged perceptron. 

In this paper, we explore more extensively and systematically two flavors of the CLTC problem: news topic classification and textual churn intent detection (TCID) in social media. In particular, \begin{inparaenum}[(1)] \item we test the hypothesis that embeddings with context are more effective, by multi-tasking the learning of multilingual word embeddings and text classification; \item we explore neural architectures for CLTC; and \item we move from bi- to multi-lingual word embeddings\end{inparaenum}. For all architectures, types of word embeddings and datasets, we notice a consistent gain trend in favor of multilingual joint training, especially for low-resourced languages.
\end{abstract}

\section{Introduction}
Text classification is one of the main applications of Natural Language Processing (NLP). However, like the majority of NLP tasks, text classification methods tend to focus on English or a handful of other languages that have text corpora of hundreds of millions of words. This is contributing to a new flavor of the digital divide: the AI divide, an inequality in the access to, use of, or impact of AI. Several technology companies are now addressing the digital divide with "next billion users" initiatives. In NLP, transfer-based approaches, such as Cross-Lingual Text Classification (CLTC) - the task of categorizing texts written in \emph{different} languages into a common taxonomy, are a promising solution.

The first CLTC studies, appearing as early as \citet{bel-03}, range from creating a single classifier for several languages by pooling the training data to training a monolingual classifier and using the translation of important terms for the other languages. Since then, the face of NLP, including CLTC, has been transformed by embeddings. Word embeddings have become a widely adopted way to transfer information from large unlabeled datasets to downstream tasks, such as sentiment analysis (\citet{acl-maas-11}), document summarization (\citet{ieee-wang-16}) or dialogue management systems (\citet{acl-zhao-16}). 

While most applications of embeddings transfer knowledge across tasks for the same language (English), multilingual embeddings aim to learn a representation common to multiple languages at the same time, making them a perfect addition to the CLTC toolbox. Indeed, a simple averaged perceptron-based CLTC is a common benchmark task to evaluate the quality of bilingual embeddings, by training CLTC on documents in a source language and testing its direct applicability to documents in a different target language.

However, the focus on CLTC as a benchmark has left several gaps. Firstly, there is no systematic comparison between CLTC with monolingual versus multilingual embeddings. Secondly, it is not clear whether and which neural architecture gives the best results for CLTC. And finally and most importantly, the multilingual embeddings are fed as such to the CLTC, treating them as universal feature representation, while recent work has shown that encoding words in context significantly improves performance in a variety of NLP tasks, for example, by transferring the encoder of a machine translation system \cite{McCann} or by multi-tasking the multilingual embeddings learning alongside with the CLTC learning.

In this paper, we address the above gaps, by establishing a comprehensive, systematic benchmarking framework \footnote{To be open-sourced after publication.} for surveying the performance of various types of embeddings on different variations of CLTC architectures. The components of the framework, corresponding to our main contributions are:
\begin{compactitem}
    \item Several \emph{CLTC architectures, adjusted to be fed directly with mono-/multi-lingual embeddings}, thus enabling mono-/multi-lingual training and the comparison between the two modes (Section \ref{pre-trained-cltc}).
    \item A representative set of state-of-the-art \emph{multilingual embeddings, obtained either via training from scratch or via offline linear projection methodologies} (e.g., Singular Value Decomposition (SVD), Canonical Correlation Analysis (CCA), Attract-Repel).
    \item A \emph{multi-tasking architecture that fine-tunes multilingual embeddings alongside the CLTC training}, thus specializing them to the CLTC task (Section \ref{spec-cltc})
\end{compactitem}
Our experiments (Section \ref{sec:results}) with two flavors of CLTC (long news stories to be classified by topics versus short tweets to be classified for churn intent) show that \textbf{the multilingual approach clearly benefits low-resource languages} and that \textbf{multilingual training outperforms language-specific models for each language}.

\section{Related Work}
In previous work, the quality of multilingual embeddings is either evaluated intrinsically by directly testing their ability to capture syntactic and semantic relationships between words. Such benchmarks include word similarity, word translation, and correlation-based evaluation. Extrinsically, those multilingual models are evaluated on their performance when used as input features to downstream semantic transfer tasks. 

One of the main downstream application of multilingual word embeddings is Cross-Lingual Document Classification benchmark (CLDC) initially defined in  \cite{coling-Klementiev-12}. They train a model on labeled documents in a source language and apply it directly to classify unlabeled documents in a target language. This aims to test the ability of multilingual embeddings to act as important agents in direct transfer learning. However, a comparison between the performance using monolingual versus multilingual embeddings is missing. \citet{acl-zhou-15} propose a methodology to learn a cross-lingual representation of sentiment information to enable sentiment classification (CLSC). They jointly train bilingual embeddings using the documents annotated with sentiments and their translations to other languages and show that the multilingual approach outperforms monolingual training.

Other work that multi-task training the multilingual embeddings with the task at hand include \cite{wang17multitask} for named entity recognition. \citet{acl-ferreira-16} propose a model that jointly learns to embed and predict classes of multilingual documents by optimizing for a loss that combines a cross-lingual training loss with a supervised document classification loss using logistic regression. Despite the simplicity of each loss component, this model manages to surpass other state-of-the-art models. The shown gain in performance in this work motivates us to investigate a multi-tasking model where a more complex model is adapted for document classification. To the best of our knowledge, we are the first to compare the gain when multilingual embeddings are used across different independent and multi-tasking architectures. 

\section{A Framework for Benchmarking Embeddings in CLTC Tasks}
In what follows, we describe the different neural network models used for CLTC at different levels of complexity and how we apply them to the multilingual setting either by directly incorporating different kinds of already trained multi-embeddings or by training the embeddings alongside with the task.

\subsection{Cross-Lingual Text Classification using Pre-trained Embeddings}
\label{pre-trained-cltc}

\begin{figure*}[!h]
\center
\subfigure[Averaged Multi-Layer Perceptron (FT-MLP) ]{\includegraphics[scale=0.6]{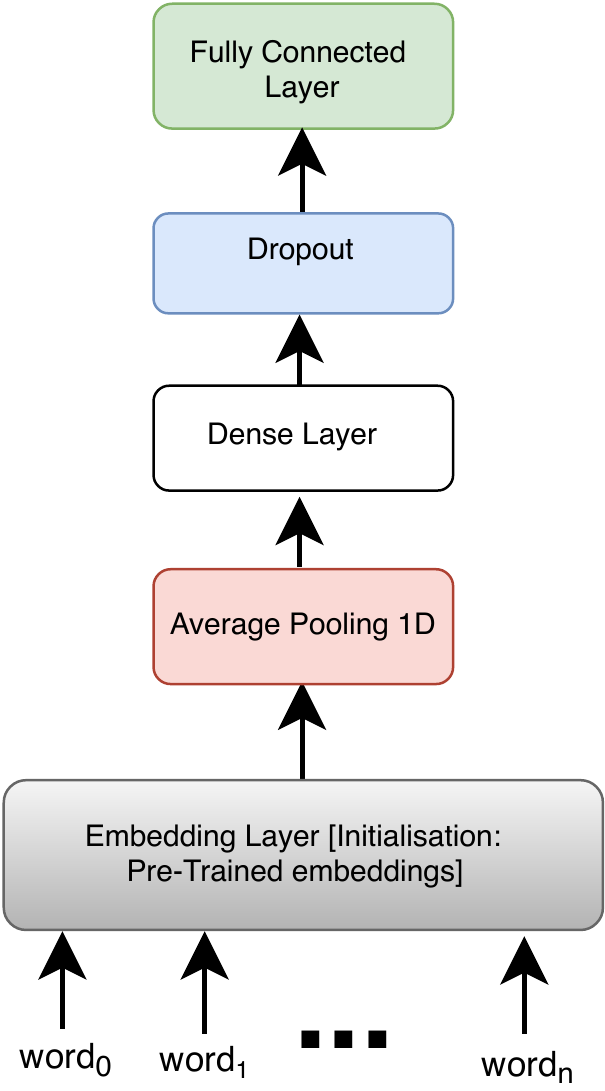}\label{ft-mlp}}
\subfigure[Multi-Filter CNN (MF-CNN)]{\includegraphics[scale=0.6]{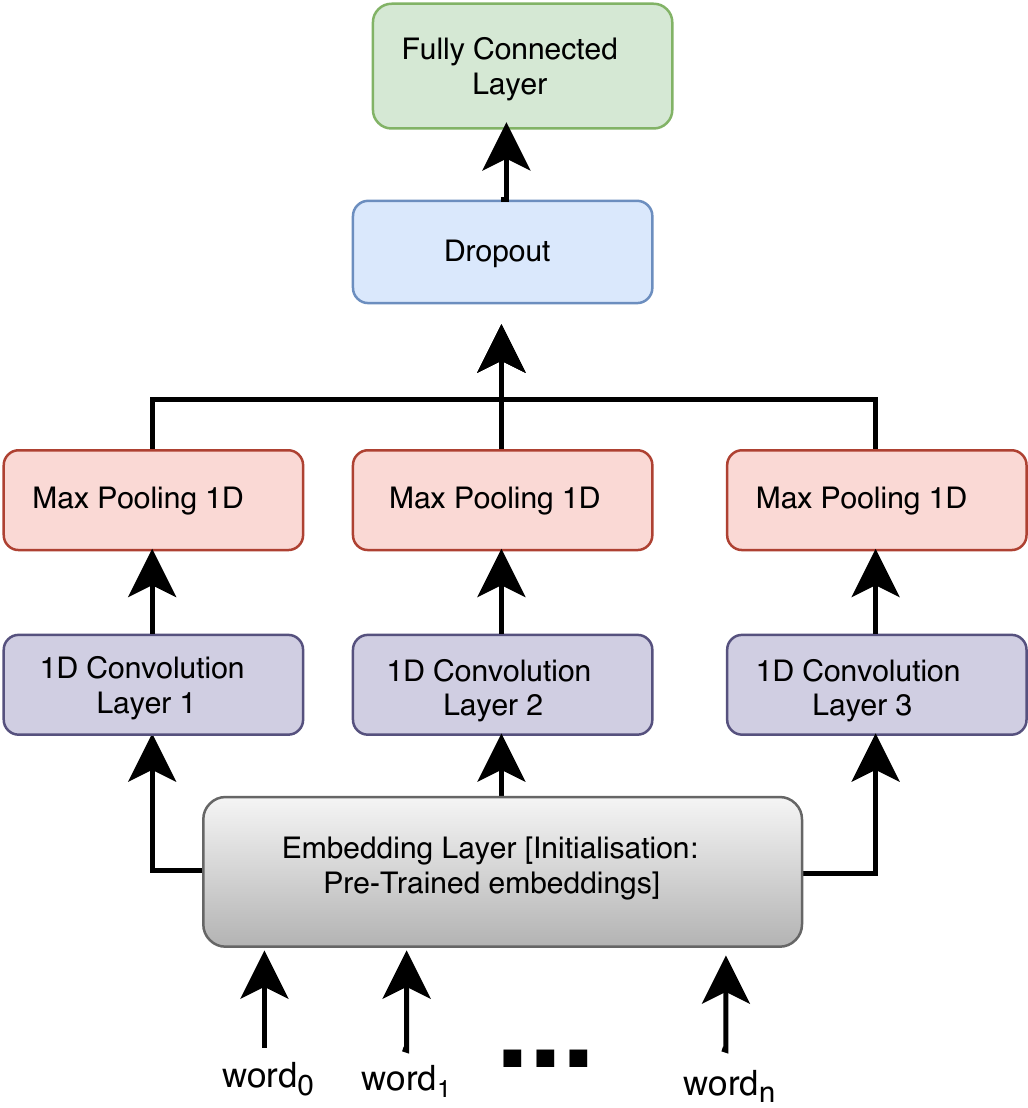}\label{fig:mf-cnn}}
\subfigure[Bidirectional GRU with Attention (bi-GRU-Att)]{\includegraphics[scale=0.6]{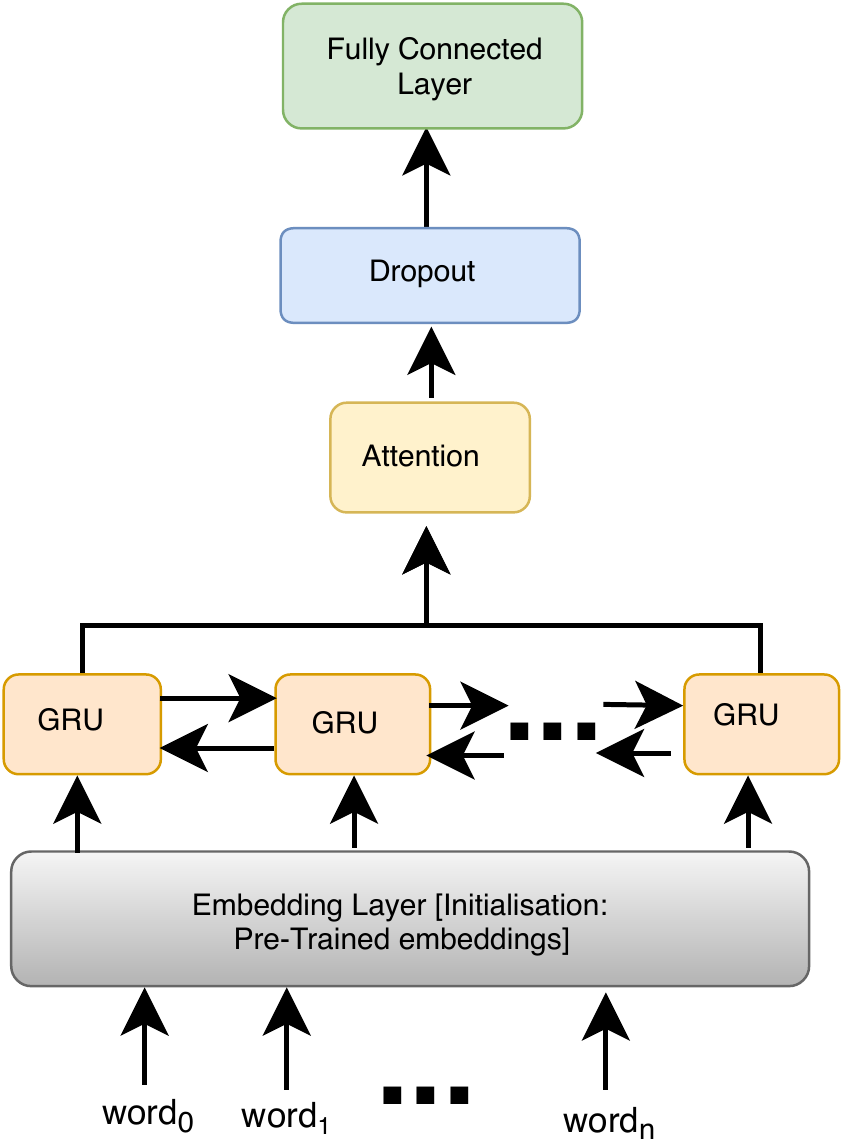}\label{fig:biGru}}
\caption{Different Document Classification Models}
\label{fig:text-class-fig}
\end{figure*}

The different variations of plain text classification models to which pre-trained embeddings are directly fed are represented in Fig. \ref{fig:text-class-fig}. In addition to fine-tuned multi-layer perceptron (FT-MLP) in Fig. \ref{ft-mlp} which is an extension of \cite{coling-Klementiev-12} averaged perceptron, we implement and evaluate other extensions namely: multi-filter convolutional neural networks and bi-directional GRU with attention. Before describing them, we explain briefly the rationale used to reproduce the set of pre-trained multilingual embeddings we work with.

\subsubsection{Pre-trained Multilingual Embeddings}
\label{embedding-method}

We obtain and reproduce several multilingual embeddings to draw fair conclusions on the potential gain of a multilingual approach applied to text classification. They have been chosen to comply with previous work proving that models with higher levels of supervision tend to perform the best \cite{empirical-comp}. We cover a  wide range of supervised methodologies to work with both models fine-tuned on top of monolingual embeddings and those trained from scratch. 

Fine-tuned multilingual embeddings models are built on top of monolingual embeddings by mapping words from different languages into one joint target space. We set English as the target space, and we learn the linear transformation that aligns other languages to English using bilingual translation pairs. We evaluate two offline fine-tuned approaches. The first variant uses Singular Value Decomposition (SVD) following the work of \citet{corr-smith-17} to produce two versions based on the type of bilingual dictionaries used: $multi(exp\_dict)$ using ground truth dictionaries and $multi(pseudo\_dict)$ using matching strings. The other variant is $multi(cca)$ which uses Canonical Correlation Analysis (CCA).

We follow Attract-Repel methodology of \citet{dblp-mrksic-17} for generating semantically specialized multilingual embeddings $multi(sem)$ by injecting monolingual and cross-lingual synonyms and antonyms as linguistic constraints to monolingual distributional vectors. We include more details on how the alignment from bilingual to multilingual is solved using VowPal Wabbit tool in Appendix \ref{app-1}.

Trained from scratch models are optimized using either cross-lingual only or both monolingual and cross-lingual constraints. The first type $multi(sen\_ali)$ follows a sentence alignment approach to optimize cross-lingual objective which consists of minimizing the distance between parallel sentences from different languages as described in Appendix \ref{app-3}. The second type of embeddings $multi(skip\_gram)$ uses skip-gram objective modified for multilingual setting as introduced by \cite{bilingual-luong-15}.

\subsubsection{Multi-Filter CNN (MF-CNN)}
We build a multi-filter CNN where convolutions of different kernel sizes are applied and concatenated as described in the work of \citet{dblp-kim-14} and shown in Fig. \ref{fig:mf-cnn}. This architecture works better than a single-filter CNN as it is shown to over-fit less.

Given an input text which consists of the concatenation of n words, an embeddings layer is used to convert the words into their corresponding m dimensional embeddings vectors $x_{1}$, $x_{2}$, ... and $x_{n}$. The input to the convolution is then the concatenation of the $nxm$ word vectors: $x_{1:n} = x_{1} \oplus x_{2} ... \oplus x_{n} $. We apply a two-dimensional convolution operation which consists of applying a filter of a window of shape: $k\times m$ where k is the number of words and m is the entire embeddings dimensionality to be traversed at a time. 

In the end, an output feature $o_{i}$ is produced from each consecutive window of $k$ words $x_{i:i+k-1}$ using the following equation: $c_{i} = f(W.x_{i:i+k-1} + b) $ where W and b are the weights and bias terms and f is a non-linearity. By applying each filter $f$ times, we obtain $f \times(n-k+1)$ feature maps. In order to concatenate different feature maps from each filter type of sizes ($k_{1}$, $k_{2}$, $k_{3}$ and so on), we apply max pooling as described in \cite{acm-collobert-11}. Then, we apply a dropout regularization to the concatenated feature before feeding the output to a dense layer with softmax activation to convert it to a probability distribution over the set of labels.

\subsubsection{Bi-directional GRU with Attention (bi-GRU-Att)}
We use a non-hierarchical version of bi-directional GRU with attention model as shown in \ref{fig:biGru}. GRU is used instead of LSTM since it more lightweight and faster to train while keeping comparable performance \cite{dblp-chung-14}. We encode the input both in its forward and backward directions to encapsulate both the past and future. On top of that, we use an attention mechanism \cite{Bahdanau:14} to get a measure of which words are more important by assigning weights of importance.

Formally, at each time step t, the GRU computes the output state as a function of the previous hidden state $h_{t-1}$ and the update gate $z_{t}$, dropping the forget gate as follows: $h_{t} = (1-z_{t}) h_{t-1} + z_{t}$. We encode each sentence $s_{i} = [x_{i0},... x_{i1}, x_{in}]$ where $x_{ij}$ are the embeddings vector for word $w_{ij}$ using GRU in the forward and backward directions: $fh_{ij} = \overrightarrow{GRU} (x_{ij})$ and $bh_{ij} = \overleftarrow{GRU} (x_{ij})$ computed for each word $w_{j}$. Those states then are concatenated to form the encoded representation for each word: $h_{ij} =[fh_{ij}, bh_{ij}]$. 

Attention weights are computed using a dense layer over all encoders' states $h_{ij}$ as shown in Eq. \ref{eq-att-1}. Then, those scores are normalized and a probability distribution is obtained using softmax as in Eq. \ref{eq-att-2}. The sentence representation is simply the weighted sum of the different encoder states by the attention weights as in Eq. \ref{eq-att-3}. 
\begin{equation}
 u_{ij} = tanh(W_{w} \times h_{ij}+b_{w})
 \label{eq-att-1}   
\end{equation}
\begin{equation}
	\alpha_{ij} = \dfrac{ exp(u_{ij}^{T} \times u_{w})}{\sum_{j=1}^{n} exp(u_{ij}^{T}\times u_{w})}
    \label{eq-att-2}
\end{equation}
\begin{equation}
    s_{i} = {\sum_{j=1}^{n} \alpha_{ij}\times h_{ij}}
    \label{eq-att-3}
\end{equation}

\noindent where $W_{w}$ and $b_{w}$ are the weights and bias of the dense layer, $u_{w}$ is the context vector that gives a high level representation of a fixed query on the words and is initialized randomly and learned during the training process. 

\subsubsection{Loss Function}
We use a weighted categorical cross entropy loss which is defined as follows:
\begin{equation}
    L_{class} = - \sum_{i=1}^{n}w_{i}\times p(i)\times log(\hat{y}_{i})
    \label{doc-cross-entro-loss}
\end{equation}
where $n$ is the number of testing instances, $w_{i}$ is the weight attributed to each instance corresponding to its class, $p(i)$ is the true label and $\hat{y}_{i}$ is the prediction. The weights are inversely proportional to the distribution of classes to circumvent the possibility of over-fitting that can be caused by an imbalanced label distribution and are computed as follows:
\begin{equation}
    w_{i} = log(\dfrac{\sum_{i=1}^{n}(max(|y_{i}|,1))}  {|y_{i}|}) + 1
\end{equation}

\subsection{Specialized Multilingual Embeddings}
\label{spec-cltc}
In addition to directly applying multilingual embeddings trained independently to text classification task, we investigate training them along with the task at hand in an end-to-end multi-tasking fashion. 

\begin{figure}[ht!]
\centering
\includegraphics[width=0.47\textwidth]{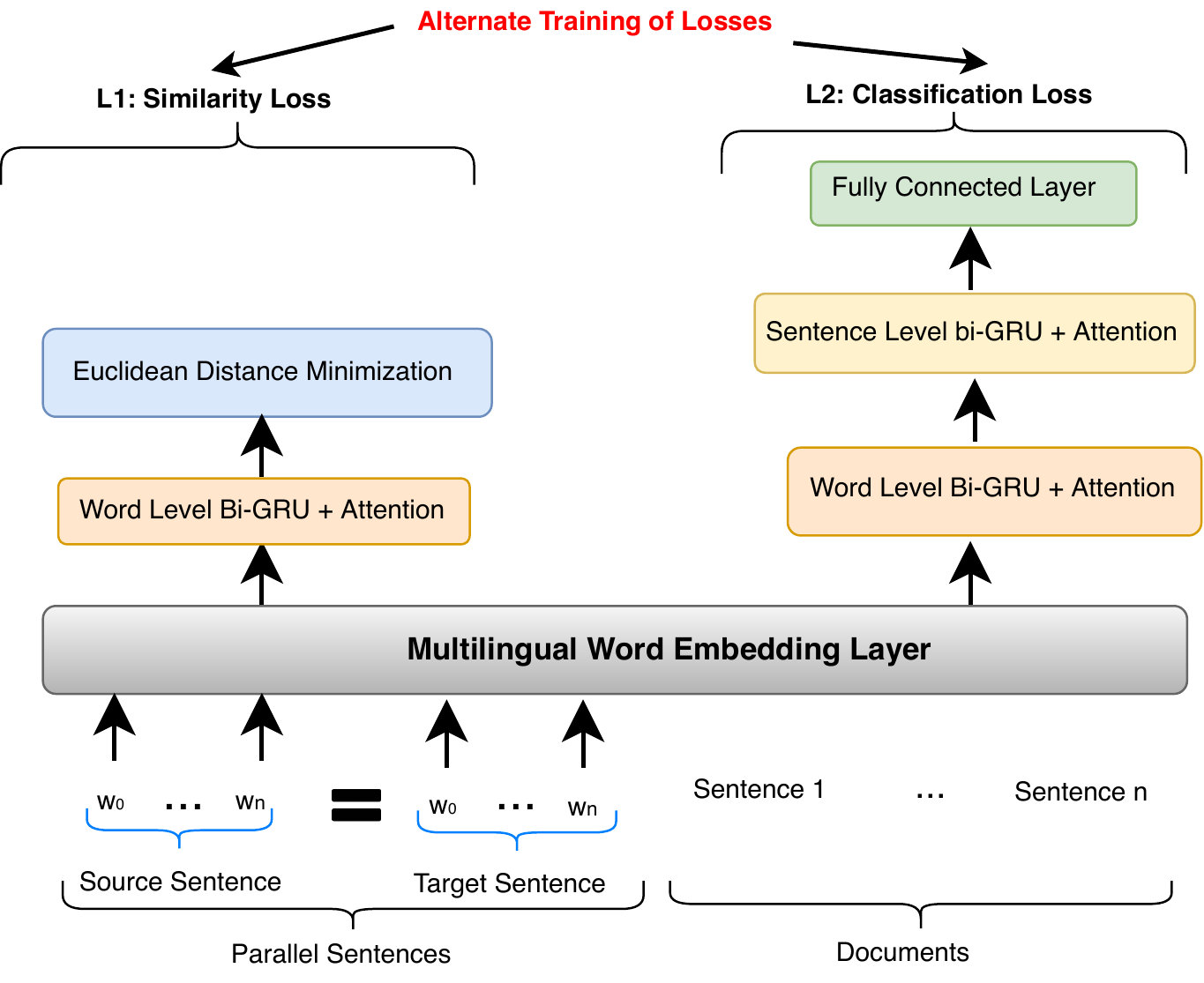}
\caption{\label{han-multi-task-fig} Multi-tasking hierarchical attention networks for CLDC and multilingual embeddings elignment}
\end{figure}

Figure \ref{han-multi-task-fig} depicts the main components of the followed architecture. The left-hand side fine-tunes multilingual embeddings using sentence alignment while the right-hand side optimizes for document classification using hierarchical bidirectional GRU attention network. The two tasks share a single embeddings layer which is tuned by the two tasks. Other layers which are shared between the tasks include word level GRU units and attention activation.

\subsubsection{Sentence Alignment (Sent-Ali)}
\label{sec:sent-alig}
The goal of this component is to construct sentence embeddings out of word embeddings using the weighted average of the output of bi-GRU states, a representation which can encapsulate word order and their importance and is more useful than taking the plain average of word embeddings. Let $S_{i}$ and $T_{i}$ be the bi-GRU encoded representation of the source and target sentences in the alignment pair ($s_{i}$, $t_{i}$) respectively. The loss $L_{sim}$ is reversely proportional to the cosine similarity between each pair ($S_{i}$, $T_{i}$) in addition to an l2-regularizer to avoid exploding gradient problem as follows:
\begin{equation}
\begin{split}
    L_{sim} = 1 - \dfrac{\sum_{i=1}^{n} S_{i}\times T_{i}}{\sqrt{\sum_{i=1}^{n}S_{i}^2}\times \sqrt{\sum_{i=1}^{n}T_{i}^2}} \\
    + \dfrac{1}{2} \times \beta \times \parallel W \parallel _{2}^{2}
\end{split}
    \label{sent-alig-loss}
\end{equation}

where $\beta$ is an arbitrarily fixed scalar that is experiment specific and $W$ is the training weights. 

\subsubsection{Hierarchical Bidirectional GRU-Attention Networks (bi-GRU-Att)}
\label{sec:doc-han}
The goal of this component is to come up with a hierarchical representation for documents (only relevant for CLDC). Unlike  \cite{naacl-yang-16}, we use a bidirectional GRU with attention at different levels. More specifically, we construct document representation using sentence encodings where each sentence representation is built from word representations where both levels of encodings use bidirectional GRUs with attention. 

\subsubsection{Learning Methodology}
\label{sec:learn-method}

We alternate between the training of the losses of the two tasks as defined in Eq. \ref{doc-cross-entro-loss} and Eq. \ref{sent-alig-loss}. Two different optimizers are adapted to each task to make the learning of one task synchronized with the other one. 

\section{Experimental Setup}
In this section, we present the approach used to compare the performance of different embeddings models and text classification architectures including datasets used for the evaluation, how experiments are designed and how models are trained.

\subsection{Datasets}
\subsubsection{Cross-Lingual Document Dataset}
\label{cldc-data}
The dataset used for CLDC is the Reuters RCV1/RCV2 corpora described in \citet{acm-lewis-04}\footnote{We obtain it under a NIST license http://trec.nist.gov/data/reuters/reuters.html}. We choose to work with this dataset since it has a sufficient amount of training instance and has been extensively used in prior research on the evaluation of multilingual embeddings which enables easy comparison with other work. RCV1 consists of about 810,000 English newswire stories, while RCV2 contains over 487,000 news stories in thirteen other languages\footnote{The thirteen languages are: Dutch, French, German, Chinese, Japanese, Russian, Portuguese, Spanish, Latin American Spanish, Italian, Danish, Norwegian, and Swedish} all made available by Reuters, Ltd. 

\begin{table}[ht]
\centering
\begin{tabular}{lllll}
\hline
                                          & \textbf{English} & \textbf{German} & \textbf{French} & \textbf{Italian} \\ \hline
\multicolumn{1}{l}{\textbf{Train}}      & 418,566          & 50,387          & 40,470          & 12,566           \\ \hline
\multicolumn{1}{l}{\textbf{Valid}} & 104,601          & 12,609          & 10,090          & 3,129            \\ \hline
\multicolumn{1}{l}{\textbf{Test}}       & 130,780          & 15,843          & 12,669          & 3,964            \\ \hline
\multicolumn{1}{l}{Total}               & 653,947          & 78,839          & 63229           & 19,659           \\ \hline
\end{tabular}
\caption{Training, Validation and Testing Distribution of RCV Dataset across Languages}
\label{rcv-stats}
\end{table}

We follow the same cross-lingual document classification benchmark defined in \cite{coling-Klementiev-12} and work on a multi-classification task with at most one single label per document among four high-level topic categories: CCAT (Corporate/Industrial), ECAT (Economics), GCAT (Government/Social), and MCAT (Markets). Table \ref{rcv-stats} shows the distribution of training, validation and testing instances per language which make up 60\%, 20\% and 20\% out of the dataset respectively.

\subsubsection{Cross-Lingual Churn Datasets}
\label{churn-data}
We use churn datasets from two languages: English and German. The English dataset \cite{aaai-amiri-15}, $EN_{T}$, contains tweets mentioning the following telecommunication brands: Verizon, AT\&T or T-Mobile. A churny tweet is one that mentions a particular brand that the Twitter user expresses an intent to leave. There are 4854 tweets in total with an annotation confidence above 0.7 out of which only 944 are churny. 
On the other hand, the German dataset \cite{conll18-christian-meryem}, $DE_{T}$, contains a total of 4339 tweets where 611 are churny regarding telecommunication operators active in German-speaking countries. 

\subsubsection{Multilingual Parallel Sentences Corpus}
We use a combination of Europarl Parallel Corpus v7.1 \cite{amt-koehn-05}, titles from Wikipedia, and parallel news commentary\footnote{http://128.2.220.95/multilingual/data/} as our sentence alignment dataset. Extracted from parliament proceedings, Europarl covers over 21 European languages. This extended corpus $PC$ is chosen because it is commonly used in the literature due to its richness and its large number of instances. The whole dataset consists of around 2.9M, 3.1M and 2.6M sentence pairs for English-German, English-French, and English-Italian. 

\subsection{Experiment Design}

For both CLDC and TCID, we design several experiments for the evaluation of different multilingual embeddings.
We train several language specific and multilingual models using different text classification architectures. In both cases, models are trained for each language independently and are used as a baseline against different multilingual embeddings models. In the end, we report only on the best and average multilingual embeddings performances in each case.

For CLDC, we evaluate three models: Fine-Tuned MLP (FT-MLP), Multi-Filter CNN (MF-CNN) and Multi-tasking embeddings with the task (HAN+Sent-Ali). For TCID, we additionally investigate the performance of bidirectional GRU with Attention (bi-GRU-Att). It was not possible to investigate the performance of bi-GRU-Att on CLDC, due to the high number of training documents and the higher number of words per document which imposes a long training time. 

\textbf{Mono:} Training on a specific language using  monolingual embeddings and testing on $EN$

\textbf{Multi:} Training on $All$ languages (i.e.  $EN+DE+FR+IT$ for CLDC and $EN+DE$ for TCID) using multilingual embeddings and testing on $EN$, $DE$, $FR$ and $IT$ for CLDC or $EN$ and $DE$ for TCID.

For all models except HAN+Sent-Ali, we run over the whole dataset. However, we only manage to run using 10K instances for HAN+Sent-Ali. This also enables us to test our hypothesis against a low data regime scenario where no language is predominant and how this impacts the gain in performance of multilingual over monolingual. On the other hand, we use the whole churn dataset in all experiments since it is already not that large.

To ensure a fair comparison between different monolingual and multilingual experiments, we use the same hyper-parameters in the design of each text classification architecture independently. In other words, we only change the parameters when switching between architectures but not when switching between monolingual and multilingual training modes. More details about the hyper-parameters used are available in Appendix \ref{app-4}. The metrics used for performance evaluation are macro F1-scores, macro precision, and macro recall.

\subsection{Pre-trained Embeddings}
Monolingual word embeddings are obtained directly from FastText \cite{acl-bojanowski-17}. Details of training $multi(exp\_dict)$ and $multi(pseudo\_dict)$ can be found in Appendix \ref{app-1}. We obtain $multi(CCA)$ and $multi(skip\_gram)$ from  \cite{corr-ammar-16} \footnote{We obtain pre-trained 512-dimensional embeddings for up to 13 languages from http://128.2.220.95/multilingual/data/.}. The linear projection from bilingual spaces to one multilingual space for $multi(sem)$ is optimized using Vowpal Wabbit tool as explained in Appendix \ref{app-2}. Details for training $multi(sent\_ali)$ can be found in Appendix \ref{app-3}. 
\section{Results}
\label{sec:results}
\begin{table*}[ht!]
\centering
\begin{tabular}{lllllllllll}
\hline
                                      &   &
                            \multicolumn{3}{c}{ \begin{tabular}[c]{@{}l@{}}\textit{\textbf{FT-}}\\\textit{\textbf{MLP}}\end{tabular}}  & 
                            \multicolumn{3}{c}{\begin{tabular}[c]{@{}l@{}}\textit{\textbf{MF-}}\\\textit{\textbf{CNN}}\end{tabular}}  &
                             \multicolumn{3}{c}{\begin{tabular}[c]{@{}l@{}}\textit{\textbf{HAN+}}\\\textit{\textbf{Sent-Ali}}\end{tabular} } 
                             \\ \hline 
                             & & F1  & P & R & F1 & P & R & F1 & P & R
                             
                             \\ \hline
\multirow{4}{*}{\textit{\textbf{EN}}} & \textit{mono}  & 91.68 & 91.62	 &   91.75              & 90.34         &  91.24	& 89.59             & 61.44    &   53.47 &  72.22                    \\
                                      & \textit{best multi} & 92.07         &  93.14     & 91.02            & 90.8       &  90.88     &   90.72             & 82.84     & 85.62     & 80.27                \\ 
                             & \textit{avg multi} &   91.62   &   91.79   & 91.45                 & 90.43    &  90.57    & 90.31               &  -  &    -  &    -
                             \\
                                      & \textit{gain}  & \textbf{0.39}            & \textbf{1.52} &  -0.73     &  \textbf{0.46}            &     -0.36  & \textbf{1.13} & \textbf{21.4}   &  \textbf{32.15}    &     \textbf{8.05}             \\ \hline
\multirow{4}{*}{\textit{\textbf{DE}}} & \textit{mono}  & 81.65  &  79.95 &	83.44	&	84.11 &	85.88 &	82.42  & 89.53     &  86.66    &  92.59                    \\
                                      & \textit{best multi} & 84.85    & 86.21 &	83.54       & 86.41  &     89.76 &	83.31   & 85.93    &   86.11    & 85.83                       \\
                            & \textit{avg multi} &   83.90 &	84.25 &	83.57 &	84.89  &	85.53  &	84.33  & - & - & -
                             \\
                                      & \textit{gain}  & \textbf{3.2}             &  \textbf{6.26} &   \textbf{0.13}   &  \textbf{2.3} & \textbf{3.88} &	\textbf{0.89} &	-3.6 &	-0.55 &	-6.76  \\ \hline
\multirow{4}{*}{\textit{\textbf{FR}}} & \textit{mono} & 81.92  &	88.44  &	76.29  &	85.77 &	88.55 &	83.17 & 76.89  &  82.22     & 72.22                    \\ 
                                      & \textit{best multi} & 88.55  &  88.55    &    88.56                 & 89.47   &  90.76    &     88.21              & 83.86      &  83.05     & 84.72                      \\
                            & \textit{avg multi} &   88.32 &	88.73 &	87.94 &		88.89  &	89.55 &	88.24   & -     &   -   & -
                             \\
                                      & \textit{gain} & \textbf{6.63} &	\textbf{0.11} & \textbf{12.27}	& \textbf{3.7} &	\textbf{2.21}	& \textbf{5.04} &	\textbf{6.97}	& \textbf{0.83}	& \textbf{12.5} \\ \hline
\multirow{4}{*}{\textit{\textbf{IT}}} & \textit{mono}  & 74.2  &	77.95  &	70.8	&	78.16 &	81.06 &	75.47  & 57.27        &   53.47    &   61.66                  \\
                                      & \textit{best multi} & 81.86  &   84.31   &  79.27                   & 81.78         &  84.67    &  79.09           & 74.72          &   73.61   &  76.11                \\
                            & \textit{avg multi} &   80.79 & 	82.77 & 	78.90	& 	80.48  & 	83.25	& 77.90  & -   &    - & - 
                             \\
                                      & \textit{gain}  &  \textbf{7.66} &	\textbf{6.36}	& \textbf{8.47} & \textbf{3.62} &	\textbf{3.61} &	\textbf{3.62} &	\textbf{17.45} &	\textbf{20.14}	& \textbf{14.45}        \\ \hline
\multicolumn{2}{l}{\textbf{Avg Gain}}                 & \textbf{4.47}	& \textbf{3.56} &	\textbf{5.03} &	\textbf{2.52} &	\textbf{2.33} & \textbf{2.67}	& \textbf{10.55}	& \textbf{13.14}	& \textbf{7.06} \\ \hline
\end{tabular}
\caption{\label{sum-cldc-tab} CLDC performance comparison between different text classification architectures highlighting gain per language}
\vspace*{-1em}
\end{table*}
\vspace*{0em}
\begin{table*}[ht!]
\hspace*{-1cm}
\begin{tabular}{llllllllllllll}
\hline
                                      &                & \multicolumn{3}{c}{\begin{tabular}[c]{@{}l@{}}\textit{\textbf{FT-}}\\\textit{\textbf{MLP}}\end{tabular}}  & 
                                  \multicolumn{3}{c}{\begin{tabular}[c]{@{}l@{}}\textit{\textbf{MF-}}\\\textit{\textbf{CNN}}\end{tabular}}  &
                                  \multicolumn{3}{c}{\begin{tabular}[c]{@{}l@{}}\textit{\textbf{bi-GRU-}}\\\textit{\textbf{Att}}\end{tabular}}   
                                   & \multicolumn{3}{c}{\begin{tabular}[c]{@{}l@{}l@{}}\textit{\textbf{bi-GRU-Att+}}\\\textit{\textbf{Sent-Ali}}\end{tabular}}  \\ \hline 
                             & & F1  & P & R & F1  & P & R & F1 & P & R & F1 & P & R \\ \hline 
\multirow{3}{*}{\textit{\textbf{EN}}} & \textit{mono}     &    68.04	 & 70.05 &	66.15	&	77.42 &	85.84 &	70.5	&	79.71 &	81.48 & 	78.01   & 52.91 & 51.26 & 55.79  \\
                                      & \textit{best} &  71.84 &	70.68 &	73.03 &  79.43	& 85.56 &	74.13  & 78.28 &	82.15 &	74.76  & 74.98 & 77.52 & 73.79                       \\
                                      & \textit{avg} &    69.68 & 	71.02 &	68.46  &		76.09 & 82.08 &	70.95 &		73.78 &	80.55 &	68.97 & - & - & -                              \\
                                      & \textit{gain}  &  \textbf{3.8} &  \textbf{0.63} &  \textbf{6.88} &  \textbf{2.01}  & -0.28 &  \textbf{3.63} &	-1.43 &  \textbf{0.67} & -3.25 &  \textbf{22.07}	&  \textbf{26.26}	&  \textbf{18}         \\ \hline
\multirow{3}{*}{\textit{\textbf{DE}}} & \textit{mono}  &  55.46 &	59.93 &	51.61	&	58.58 &	68.22 &	51.33	&	58.58 &	68.22 &	51.33  & 63.24  & 63.59 & 64.16    \\  
                                      & \textit{best} &  65.68	& 66.01 &	65.37 & 69.03 &	77.32 &	62.34 & 69.87 &	74.81 &	65.54 & 73.44 & 80.05 & 69.21             \\
                                      & \textit{avg} &   64.05 &	66.07 &	62.20 &		66.93 &	73.98 &	61.14 &	 66.34	& 72.84 &	60.94  & - & - & -         \\
                                      & \textit{gain} &  \textbf{10.22}	&  \textbf{6.08}	&  \textbf{13.76}	&  \textbf{10.45}	&  \textbf{9.1} &  \textbf{11.01} &  \textbf{11.29}	&  \textbf{6.59}	&  \textbf{14.21}	&  \textbf{10.2}	&  \textbf{16.46}	&  \textbf{5.05}     \\ \hline
\multicolumn{2}{l}{\textbf{Avg Gain}}                 &  \textbf{7.01}	&  \textbf{3.35}	&  \textbf{10.32}	&  \textbf{6.23}	&  \textbf{4.41}	&  \textbf{7.32}	&  \textbf{4.93}	&  \textbf{3.63}	&  \textbf{5.48}	&  \textbf{16.13}	&  \textbf{21.36}	&  \textbf{11.52}               \\ \hline
\end{tabular}

\caption{\label{sum-churn-tab} TCID performance comparison between different text classification architectures highlighting gain per language}
\end{table*}
Table \ref{sum-cldc-tab} summarizes F1-score, precision and recall performance for CLDC task by comparing the different gains of the best multilingual over monolingual training for each text classification architecture. In general, for all languages and all text classification architectures, multilingual training wins over monolingual training with an average improvement in F1-score of $4.47\%$ and $2.52\%$ and $10.55\%$ for FT-MLP and MF-CNN and HAN+Sent-Ali respectively. 

After examining different multilingual embeddings (Appendix \ref{app-5-1} and  \ref{app-5-2}), we notice that the gain is more or less the same and not as significant as the fluctuations in gains when changing text classification architecture. For these reasons, we report only best performant multilingual model in each case, and we average over all multilingual embeddings for a more concise and clear analysis. 

Using FT-MLP, the improvement is well pronounced mostly for Italian (the most resource scarce language)  with an increase of $7.66\%$ in F1-score followed by French and German with increases of $6.63\%$ and $3.2\%$ respectively which matches the order of languages in terms of the number of training and validation instances according to Table \ref{rcv-stats}. This finding is similar to MF-CNN and confirms our hypothesis that the less resourced a language is, the more likely it is to benefit from multilingual training. Although there is a gain in performance for English, it is marginal for both architectures (only $0.36 $ at most). Obtaining a monolingual performance for English always on par with multilingual performance is not at all surprising as English is the dominant language accounting for more than $80\%$ of the training and validation data. 

We notice that MF-CNN performs slightly better than FT-MLP with an across language average gain in performance of $2.23\%$ and has a lower gap between multilingual and monolingual, which is not counter-intuitive since MF-CNN is more complex than FT-MLP and it has more parameters to train which leads even monolingual models to converge better. This verifies the hypothesis that the gain that comes from multilingual aggregation is more pronounced the more shallow the model is.  

On the other hand, performance when multi-Tasking embeddings training alongside with the classification task using HAN architecture is even lower compared to other shallower models like MF-CNN and FT-MLP. This can be explained by the low data regime adopted. The results support our assumption by showing a more significant gain of multilingual over monolingual since all languages are low resourced in this case. For example, higher gain in English in case of HAN+Sent-Ali compared to other models ($21.4\%$ for HAN+Sent-Ali versus $0.46\%$ and $0.39\%$ for FT-MLP and MF-CNN respectively) is due to the fact that English, in this model, is treated as a low-resourced language as maximum of 10K instances from each language are used for training. 

Table \ref{sum-churn-tab} compares between results obtained for TCID using different text classification architectures across different training modes and embeddings. The results show that in general multilingual models tend to outperform monolingual baselines for both English and German irrespective of the embeddings model used with an average increase of $7.01\%$, $6.23\%$ and $4.93\%$ FT-MLP, MF-CNN, and bi-GRU-Att respectively. 

We notice that the gap between multilingual and monolingual becomes smaller: $7.01\%$, $6.23\%$, and $4.93\%$ for the three architectures from less to more complex which matches our previous finding in CLDC. The difference between the degree of improvement of multilingual versus monolingual for English and German is due to the fact that English dataset has already what it takes to learn classification patterns while German benefits more from the aggregation of more languages to learn complex patterns that are not present in German alone. In all cases, we notice that multilingual embeddings performance are close to each other with an average of $67.1 \pm 0.86$ and far from monolingual performance for FT-MLP for example. Appendix \ref{app-5-1} and  \ref{app-5-2} provide a fine-grained analysis over all multilingual embeddings for CLDC and TCID tasks respectively.

\section{Conclusion and Future Work}
In this paper, we put in place a systematic multi-dimensional comparative analysis of multilingual embeddings on two variations of Cross-Lingual Text Classification (CLTC) tasks. Our approach has the advantage of being unified for training across languages leveraging different multilingual embeddings methods and an end-to-end benchmark for their evaluation against their monolingual counterparts. The embeddings covered in our analysis span a diverse spectrum of methodologies covering those fine-tuned on top of monolingual embeddings, those trained from scratch, and those learned jointly with the task. We test both in an imbalanced data scenario with English being the most dominant language and in a low data regime and witnessed a consistent gain of multilingual approach especially for low-resource languages for all text classification architectures and for both datasets. 

Although this study focuses on four languages at most: English, French, German and Italian, the described models and evaluation strategy can be extended to more languages. Testing more languages especially under-resourced ones can be explored in future work. It is also worth investigating ways of making multi-tasking architecture scalable to test it on the whole imbalanced document dataset.

\bibliography{naaclhlt2019}
\bibliographystyle{acl_natbib}

\appendix

\section{Training Offline Embeddings}
\label{app-1}
We build multilingual embeddings which map words from different languages into one joint vector space by learning translations of monolingual embeddings into a target space. We set English as the target space and we learn the transformation matrix that aligns other languages to English using bilingual translation pairs. In other words, this approach fine-tunes non-English embedding by applying a linear transformation that maps them into the English space. 

We learn the alignment on top of monolingual embeddings using the training split of the expert bilingual dictionary where the problem of building bilingual embeddings reduces to learning the linear transformation matrix $W_{\ell \rightarrow{EN}}$ which maps the source $\ell$ monolingual space into the English space where $\ell \in L-{EN}$. Formally, given X and Y monolingual word vector matrices for the source and target spaces, the goal is to learn $W_{\ell \rightarrow{EN}}$ that maximizes the cosine similarity defined by:
\begin{equation}
    \max_{W_{\ell \rightarrow{EN}}} \sum_{i=1}^{n} {y_{i}^{T}} W x_{i}
\end{equation}

\citet{corr-smith-17} proves that this optimization objective can be solved directly and efficiently using SVD of the product of the paired dictionary matrices: 
\begin{equation}
    M = Y^{T}_{D} \cdot X_{D}= U\cdot \sum \cdot V^{T}
\end{equation} 

The resulting U and V vectors are orthonormal matrices whose product gives us the desired transformation matrix $W_{\ell \rightarrow{EN}}$. We also apply dimensionality reduction by keeping only the first rows in matrices U and V which correspond to large values in the diagonal matrix $\sum$. 

We train two variants of this approach: $multi(exp\_dict)$ and $multi(pseudo\_dict)$. For training $multi(exp\_dict)$, we use ground truth bilingual dictionaries as introduced in \cite{corr-conneau-17}\footnote{A large repository of up to 110 bilingual dictionaries covering high and low resource languages is available in https://github.com/facebookresearch/MUSE} consisting of translation pairs for each pair of source and target languages (where the target language is always English). Only the train split (consisting of 5000 pairs) is used for training while 1500 pairs are used for testing the quality of the embeddings before feeding them to the downstream applications. For both $multi(exp\_dict)$ and $multi(pseudo\_dict)$, we use dimensionality reduction on top of SVD by considering only the first significant rows corresponding to a value threshold of 1 in the diagonal vector.
\section{Training of Semantic Specialized Multilingual Embeddings using using Vowpal Wabbit}
\label{app-2}

To learn the alignment from bilingual to multilingual space, we learn the weights for two linear projections: from EN-FR to EN-DE and from EN-IT to EN-DE to bring the French part of EN-FR and Italian part of EN-IT to the same joint space as EN-DE. We solve each linear projection using logistic regression optimized using stochastic gradient descent. 

Here, we describe the approach for learning the mapping $W^{*} = W_{EN-FR \rightarrow{EN-DE}}$. The idea is to make use of the inherent parallelism between the two spaces in the sense that English vectors for words in space EN-FR should be aligned to vectors of the same words in space EN-DE. Formally, let $u_{i}$ and $v_{i}$ be the vectors of word i in space EN-FR and EN-DE respectively. So, we learn the matrix $W^{*}$ such that $u_i 	\simeq W^{*} v_{i}$. This can be solved by minimizing the Euclidean distance between English words shared between the two spaces as follows:
\begin{equation}
        \sum_{i=1}^{n} \parallel u_{i} - W^{*}\cdot v_{i}\parallel^{2} = \parallel U-W^{*}\cdot V \parallel ^{2}
\end{equation}
where U and V are embeddings matrices where each row corresponds to vector in EN-FR and EN-DE of each word shared between the two spaces and $\parallel.\parallel_{F}$ is the Frobenius norm.

To solve this m-variate linear regression, we use stochastic gradient descent (SGD) which is solved using Vowpal Wabbit, a library that can handle large-scale data efficiently. To comply with VW inability to deal with multidimensional output, we split the problem to single output linear regression sub-problems. Therefore, for each sub-problem, we create a VW file for each embeddings dimension $j=[1,2, .., n]$ of $u$. The format of the file looks like:  

\begin{center}
$u_{1j} | 1:v_{11} 2:v_{12} ... n:v_{1m}$
\end{center}

\begin{center}
$u_{2j} | 1:v_{21} 2:v_{22} ... n:v_{2m}$
\end{center}

\begin{center}
$u_{nj} | 1:v_{n1} 2:v_{n2} ... n:v_{nm}$
\end{center}
where n is the number of words, m is the dimensionality of the embeddings. Running optimization for this file results in the $j^{th}$ column of the desired transformation $W^{*}$. In the end, this transformation is applied to French vectors in EN-FR (and Italian vectors in EN-IT with the same methodology) leaving German and English vectors of EN-DE unchanged. We run for 100 passes and Vowpal Wabbit fines tunes by itself the learning parameters.

\section{Training Sentence Alignment}
\label{app-3}
\paragraph{Bilingual Case}
Training embeddings by optimizing the cross-lingual objective using sentence alignment means to train a model that maximizes the semantic similarity between parallel sentences. Formally, given pairs of parallel sentences in two languages $l_{1}$ and $l_{2}$, the goal is to find the embeddings matrices $P$ and $Q$ which transform sentences in $l_{1}$ and $l_{2}$ to one common space. For that purpose, we minimize the sum of the distances between the embeddings representation of aligned sentences as follows:

\begin{equation}
	\begin{split}
        L = \dfrac{1}{2\times N} \times \sum_{i=0}^{N}\parallel P - P_{0} \parallel^{2} + \\
         \dfrac{1}{N}\times \mu \times \sum_{i=0}^{N}\parallel P^{T}s_{i} - Q^{T}t_{i} \parallel_{1} + \\
          \dfrac{\mu_{S}}{2}\times \parallel P \parallel^{2}_{F} + \dfrac{\mu_{T}}{2}\times \parallel Q \parallel^{2}_{F}
    \end{split}
\end{equation}
where $N$ is the total number of aligned sentences, $s_{i} \in l_{1}$, $t_{i} \in l_{2}$ and $(s_{i}, t_{i}) \in PC$ and $\mu$, $\mu_{S}$, $\mu_{T}$ are regularization terms. Here l1-distance was chosen instead of l2-distance for its robustness against outliers. 

We take advantage of monolingual embeddings to initialize $P$ with $P_{0}$. $P$ and $Q$ are optimized using gradient descent with steps $P = step_{P} \times \delta(P)$ and $Q = step_{Q} \times \delta(Q)$ to optimize $P$ and $Q$ respectively as follows:
\begin{equation}
        step_{P}= \dfrac{\eta} {\epsilon+ \sqrt{\parallel \delta(P) \parallel^{2} }}
\end{equation}
\begin{equation}
        step_{Q}= \dfrac{\eta} {\epsilon+ \sqrt{\parallel \delta(Q) \parallel^{2} }}
\end{equation}
where the gradients are computed as follows: 
\begin{equation}
	\begin{split}
        \delta(P) = \dfrac{\mu}{N} \times S \cdot T \cdot \parallel P^{T}s_{i} - Q^{T}t_{i} \parallel_{1} + \\
         (\dfrac{1}{|P|}+\mu_{s}) \times P - \dfrac{P_{0}}{|P|}
    \end{split}
\end{equation}

\begin{equation}
	\begin{split}
        \delta(Q) = - \dfrac{\mu}{N} \times T \cdot T \cdot \parallel P^{T}s_{i} - \\ 
        Q^{T}t_{i} \parallel_{1} + \mu_{t} \times Q
    \end{split}
\end{equation}
The list of parameters used for our experiment to generate embeddings is as detailed in table \ref{sent-parameters}.

\begin{table}[ht]
\centering
\begin{tabular}{|l|l|}
            \hline
            \textbf{Param} & \textbf{Val} \\ \hline
            $\mu$       & 1e-9              \\ \hline
            $\mu_{s}$        & 1e-11             \\ \hline
            $\mu_{t}$            & 1e-11                      \\ \hline
            num epochs          & 50         \\ \hline
            
            $\eta$          & 1         \\ \hline
            
            $\epsilon$          & 1e-12         \\ \hline
            
            Dimension          & 300         \\ \hline
            \begin{tabular}[c]{@{}l@{}}Learning \\ Rate \end{tabular}           & 10-2           \\ \hline Batch size           & 64             \\ \hline
        \end{tabular}
\caption{Training Parameters for Sentence Alignment}
\label{sent-parameters}
\end{table}

\paragraph{Multilingual Extension}
The multilingual extension is straightforward as the bilingual objective function is additive. Therefore, the multilingual objective consists of the sum of multiple bilingual objectives which is equivalent to one bilingual objective where the source language for sentences is any non-English language, and the target is English. Thus, we train multilingual embeddings using a concatenation of all sentences from German, French, and Italian to learn $P$ and English sentences to learn $Q$.

\section{Implementation and Hyperparameter Choices}
\label{app-4}
\begin{table*}[ht!]
    \begin{minipage}{.2\linewidth}
         \begin{tabular}{ll}
            \hline
            \textbf{Param} & \textbf{Val} \\ \hline
            \begin{tabular}[c]{@{}l@{}}Dense \\ Units L1 \end{tabular}        & 512              \\ \hline
            \begin{tabular}[c]{@{}l@{}}Dense \\ Act L1 \end{tabular}        & relu              \\ \hline
            Dropout            & 0.7           \\ \hline
            Optim          & Ada             \\ \hline \begin{tabular}[c]{@{}l@{}}Learning \\ Rate \end{tabular}           & 10-2 
            
            \\ \hline
           Patience           & 20           \\ \hline
           Batch           & 64          \\ \hline
            
    \end{tabular}
    \\\\\\ \centering a) FT-MLP
    \end{minipage}
    \begin{minipage}{.2\linewidth}
        \begin{tabular}{ll}
            \hline
            \textbf{Param} & \textbf{Val} \\ \hline
             \begin{tabular}[c]{@{}l@{}}Kernel \\ Sizes \end{tabular}        & 3,4,5              \\ \hline
            \# Filters         & 200            \\ \hline
            Dropout            & 0.3           \\ \hline
            Optim          & \begin{tabular}[c]{@{}l@{}}Ada\\\end{tabular}           \\ \hline
            \begin{tabular}[c]{@{}l@{}}Learning \\ Rate \end{tabular}           & 10-3 
            
            \\ \hline
            Patience           & 20             \\ \hline
            Batch           & 64        \\ \hline
        \end{tabular}
       \\\\ \\b) MF-CNN
    \end{minipage} 
    \begin{minipage}{.18\linewidth}
        \begin{tabular}{ll}
            \hline
            \textbf{Param} & \textbf{Val} \\ \hline
            \begin{tabular}[c]{@{}l@{}}\# GRU \\ units\end{tabular}       & 150            \\ \hline
            \begin{tabular}[c]{@{}l@{}}GRU \\ activation\end{tabular}       & tanh            \\ \hline
            Dropout            & 0.3            \\ \hline
            Optim          & \begin{tabular}[c]{@{}l@{}}Ada\\\end{tabular}           \\ \hline
            \begin{tabular}[c]{@{}l@{}}Learning \\ Rate \end{tabular}           & 10-3 
            
            \\ \hline
            Patience           & 20             \\ \hline
            Batch           & 64        \\ \hline
        \end{tabular}
        \\ \\c) bi-GRU-Att
    \end{minipage} 
        \begin{minipage}{.2\linewidth}
        \begin{tabular}{ll}
            \hline
            \textbf{Param} & \textbf{Val} \\ \hline
            \begin{tabular}[c]{@{}l@{}}\# GRU \\ units \end{tabular}       & 50            \\ \hline
            \begin{tabular}[c]{@{}l@{}}GRU \\ activation \end{tabular}       & tanh            \\ \hline
            Dropout            & 0.5                    \\ \hline
            \begin{tabular}[c]{@{}l@{}}Optim\\Task 1\end{tabular}         & \begin{tabular}[c]{@{}l@{}}Ada\\(10-3)\end{tabular}            \\
            \hline
           \begin{tabular}[c]{@{}l@{}}Optim\\Task 2\end{tabular}               & \begin{tabular}[c]{@{}l@{}}Ada\\(10-2)\end{tabular}            \\
            \hline beta          & 1e-10             \\ \hline Batch           & 15             \\ \hline
        \end{tabular}
        \\d) Multi-Tasking
    \end{minipage} 
    \begin{minipage} {.2\linewidth}
    \begin{tabular}{ll}
        \hline
        \textbf{Param} & \textbf{Val} \\ \hline
        $\mu$       & 1e-9              \\ \hline
        $\mu_{s}$        & 1e-11             \\ \hline
        $\mu_{t}$            & 1e-11                      \\ \hline
        \# epochs          & 50         \\ \hline
        
        $\eta$          & 1         \\ \hline
        
        $\epsilon$          & 1e-12         \\ \hline
        \begin{tabular}[c]{@{}l@{}}Learning \\ Rate \end{tabular}           & 10-2           \\ \hline Batch           & 64             \\ \hline
    \end{tabular}
    \\ \\ \\ \\ e) Sentence Alignment
    \label{sent-ali-parameters}
    \end{minipage}

    \caption{\label{hyper-tab} HyperParameters for different text classification architectures and sentence alignment}
\end{table*}

\begin{table*}[ht!]
\centering
\begin{tabular}{lllllllll}
\hline               \textbf{Train}                          & \textbf{Test}                  & \textbf{Embeddings}          & \multicolumn{3}{c}{\textbf{FT-MLP}} & \multicolumn{3}{c}{\textbf{MF-CNN}} \\ \hline  & &                         & F1 & P & R & F1 & P & R    \\ \cline{4-9}
\textbf{EN}                           & \multirow{7}{*}{\textbf{EN}} & \textit{mono}                & 91.68                                                    & 91.62                                                             & 91.75     & 90.34                                                             & 91.24                                                             & 89.59                                                 \\ \multirow{6}{*}{\textbf{All}}
                                 &                       & \textit{multi(pseudo\_dict)} & 91.46                                                             & 91.21                                                             & 91.71  & 90.46                                                   & 90.42                                                    & 90.52                                                           \\
                                 &                                         &   \textit{multi(exp\_dict)} &  91.61                                                   &      91.40                                              &       \textbf{91.81}  &  90.34                                                   &      90.45                                              &       90.26                  \\
                                   &                                         &                                                   \textit{multi(CCA)}        & 91.48                                                             & 91.91                                                    & 91.05      &  \textbf{90.8}                                                             & \textbf{90.88}                                                             & 90.72                                                                              \\
                                   &                                         &                                                  \textit{multi(sem)}        & \textbf{92.07}                                                             & \textbf{93.14}                                                    & 91.02  &           90.14                                         &  90.85                                                   &         89.48                               \\
                                 &                                         &                                     \textit{multi(sent\_ali)}        & 91.61                                                             & 91.68                                                    & 91.55    &                  90.18                                   &                   90.13                                 &      90.24              
                               \\ &                                         &                                                    \textit{multi(skip\_gram)}        & 91.49                                                             & 91.4                                                   & 91.58        &  90.66                                                            &  90.69                                                   & \textbf{90.64}                              \\\cline{1-9}
                                 
                                 \textbf{DE}                           & \multirow{7}{*}{\textbf{DE}} & \textit{mono}                & 81.65                                                             & 79.95                                                             & 83.44                                         & 84.11                                                             & 85.88                                                             & 82.42                     \\
                                  \multirow{6}{*}{\textbf{All}} &    &                            \textit{multi(pseudo\_dict)} & 84.44                                                    & 85.66                                                    & 83.25  & 83.77                                                    & 84.64                                                    & 82.97                                                      \\
                                 &                                         &                       
                                 \textit{multi(exp\_dict)} &  \textbf{84.85}                                                  &      \textbf{86.21}                                              &       83.54       &               86.37                                      &                          83.91                          &         \textbf{88.97}                                                                             \\ &                                         &                                                 \textit{multi(CCA)}        & 83.07                                                            & 82.68                                                    & 83.46  & 84.46                                                             & 85.5                                                            & 83.45                                    \\
                                   &                                         &                          
                                 \textit{multi(sem)}        & 83.15                                              & 83.13                                                     & 83.19                   &           83.79                                          &  84.13                                                   &        83.46             \\
                                 &                                         &                        \textit{multi(sent\_ali)}        & 83.93                                                             & 83.46                                                             & \textbf{84.42}                      &     \textbf{86.41}                                                &                             \textbf{89.76}                      &     83.31                                    \\ &    &                                                                          \textit{multi(skip\_gram)}        &    83.96                                                          &    84.35                                                 & 83.57      &   84.52                                                           &   85.21                                                  & 83.84                              \\\cline{1-9}
                                  \textbf{FR}                           & \multirow{7}{*}{\textbf{FR}} & mono                         & 81.92                                                             & 88.44                                                             & 76.29                                       & 85.77                                                             & 88.55                                                             & 83.17                    \\
                                 \multirow{6}{*}{\textbf{All}}   &                          & \textit{multi(pseudo\_dict)}          & 88.51                                                             & 89.54                                                             & 87.5                             & 88.69                                                   & 88.72                                                    & 88.66                                      \\
                                 &                                         &                     \textit{multi(exp\_dict)} &  88.27                                                   &      \textbf{89.99}                                              &       86.62                               &   88.03                                                  &       88.83                                             &          87.25            \\  &                                         &                                               \textit{multi(CCA)}        &    88.34                                                          &  88.38                                                   &   88.31                       & \textbf{89.47}                                                             & \textbf{90.76}                                                            & 88.21          \\
                                   &                                         &          
                                 \textit{multi(sem)}        & 87.75                                                            &  86.97                                                   & 88.55          &    88.55                                                 &                                88.85                    &     88.26                   \\
                                 &                                         &                       \textit{multi(sent\_ali)}                 & \textbf{88.55}                                                    & 88.55                                                    & \textbf{88.56}   &         89.43                                     &                                90.11                    &           \textbf{88.75}                                                                      
                                          \\ &                                         &                                            \textit{multi(skip\_gram)}        & 88.52                                                             & 88.97                                                    & 88.07          &  89.16                                                             &  90.04                                                    & 88.29                             \\\cline{1-9}
                                  \textbf{IT}                           & \multirow{7}{*}{\textbf{IT}} & mono                         & 74.2                                                             & 77.95                                                    & 70.8                                                     & 78.16                                                             & 81.06                                                             & 75.47          \\
                                 \multirow{6}{*}{\textbf{All}} &  &                          \textit{multi(pseudo\_dict)}          & \textbf{81.86}                                                    & \textbf{84.31}                                                             & 79.27                           & 80.11                                                 & 83.41                                                    & 77.07                       \\
                                 &                                         &                         \textit{multi(exp\_dict)} &  80.76                                                   &      84.15                                              &       77.65                             &       78.56                                              &    81.40                                                &        75.92                    \\  &                                         &                                              \textit{multi(CCA)}        &    81.53                                                          & 81.17                                                     &   81.89                     & \textbf{81.78}                                                            & \textbf{84.67}                                                             & 79.09           \\
                                   &                                         &                      
                        \textit{multi(sem)}        & 80.82                                                             & 82.96                                                    & 78.80               &            80.76                                         &                                   81.81                 &      \textbf{79.74}              \\
                                 &                                         &                       
                        \textit{multi(sent\_ali)}                 & 78.98                                                            & 82.37                                                             & 75.87   &         80.18                                           &                           84.14                         &   76.57 
                                                    
                                 \\ &                                         &                                 \textit{multi(skip\_gram)}        &  80.76                                                            &  81.64                                                   & \textbf{79.89}               & 81.49                                                             & 84.07                                                    &  79.07                         \\\cline{1-9}
\end{tabular}
\caption{\label{cldc-all-tab} CLDC Performance Comparison between different training modes with different embeddings using different text classification architectures}
\end{table*}

\begin{table*}[ht!]
\centering
\begin{tabular}{llllllllllll}
\hline
\textbf{Train}                          & \textbf{Test}                  & \textbf{Embeddings}          & \multicolumn{3}{c}{\textbf{FT-MLP}} & \multicolumn{3}{c}{\textbf{MF-CNN}} & \multicolumn{3}{c}{\textbf{bi-GRU-Att}} \\ \hline  & &                         & F1 & P & R & F1 & P & R & F1 & P & R    \\ \cline{4-12}
 \textbf{EN}                       & \multirow{7}{*}{\textbf{EN}} & \textit{mono}                & 68.04                                                             & 70.05                                                             & 66.15    & 77.42                                                            & 85.84                                                    & 70.5                                        & \textbf{79.71}                                                             & 81.48                                                             & \textbf{78.01}                 \\
                      \multirow{6}{*}{\textbf{All}}   &                                 & \textit{multi(pseudo\_dict)} & \textbf{71.84}                                                             & 70.68                                                             & \textbf{73.03}    & 79.31                                                     & 83.51                                                             & \textbf{75.51}   & 76.86                                                    & 77.73                                                             & 76.02                                                               \\
                     &                                                                           & \textit{multi(exp\_dict)} & 67.12                                                             & 69.22                                                            & 65.14  &                   \textbf{79.43}                                          &        \textbf{85.56}                                                 &  74.13   & 78.28                                                    & 82.15                                                             & 74.76                                                       \\
                    &                                                                           & \textit{multi(CCA)}          & 70.89                                                   & 70.52                                                    & 71.28              & 76.76                                                             & 82.52                                                             & 71.76    & 78.19                                                             & 82.42                                                             & 74.39                                                \\
                     &                                                                           &
                     \textit{multi(sem)}          & 68.12                                                             & 70.80                                                             & 65.64    & 73.37                                                             & 80.0                                                              & 67.76    & 73.53                                                            & 79.34                                                             & 68.51                                                                               \\
                     &                                                                         & \textit{multi(sent\_ali)}        & 69.19                                                             & \textbf{72.55}                                                             & 66.14    & 69.45                                                             & 74.55                                                             & 65.01    & 69.39                                                            & 71.23                                                             & 67.65                                                                 \\
                     &                                                                         & \textit{multi(skip\_gram)}        & 70.91                                                           & 72.35                                                             & 69.52     & 78.22                                                             & 86.33                                                             & 71.5    & 66.43                                                             & \textbf{90.44}                                                             & 52.5                                                                                                                                          \\
                      \cline{1-12} 
                     \textbf{DE}                       & \multirow{7}{*}{\textbf{DE}} & \textit{mono}                & 55.46                                                             & 59.93                                                             & 51.61      & 58.58                                                             & 68.22                                                              & 51.33      & 58.58                                                             & 68.22                                                             & 51.33                                                                                                                                                     \\ 
                     \multirow{6}{*}{\textbf{All}} &                                 & \textit{multi(pseudo\_dict)} & 64.08                                                             & 65.38                                                             & 62.83        & \textbf{69.03}                                                    & \textbf{77.32}                                                    & \textbf{62.34}    & \textbf{69.87}                                                 & 74.81                                                    & \textbf{65.54}                                                                                                                \\ 
                     &                                         &            
                     \textit{multi(exp\_dict)} & 62.13                                                             & 66.53                                                            & 58.27        &           66.81                                                  & 74.40                                                        &  60.62       &  68.71                                                            &    \textbf{75.74}                                                       &  62.87       \\ 
                     &                                                          & \textit{multi(CCA)}          & 65.02                                                             & 65.90                                                             & 64.16          & 64.45                                                             & 72.97                                                             & 57.71       & 65.39                                                             & 74.12                                                             & 58.50                                                                                            \\ &                                                               &
                     \textit{multi(sem)}          & 64.15                                                             & 65.99                                                             & 62.42    & 68.46                                                             & 76.1                                                              & 62.21       & 66.26                                                             & 73.24                                                             & 60.5                                                 \\ 
                                            &                                    & \textit{multi(sent\_ali)}        & \textbf{65.68}                                                    & 66.01                                                    & \textbf{65.37}    & 65.57                                                             & 70.29                                                             & 61.45       & 67.17                                                            & 71.89                                                             & 63.03                                                                                                                  \\
                     &                                                                         & \textit{multi(skip\_gram)}        & 63.21                                                          & \textbf{66.6}                                                             & 60.14    & 67.25                                                             & 72.79                                                             & 62.5  & 60.63                                                           & 67.24                                                             & 55.2                                                \\ \hline
\end{tabular}
\caption{\label{churn-all-tab}Comparison of Detection Results using different text classification architectures}
\end{table*}
Table \ref{hyper-tab} shows the different hyperparameters used for each model. For FT-MLP, we use a first dense layer with 512 units and rectified linear unit activation prior to the second dense layer that directly precedes softmax activation, a dropout layer of 0.7, an Adam optimizer with learning rate 10-2. For MF-CNN, we use 3 types of filters with kernel sizes 3, 4 and 5 consisting of 200 filters each, a dropout of 0.3 and Adam optimizer with learning rate 10-3. bi-GRU-Att uses 150 GRU units with tanh as an activation function, dropout layer of 0.3 and Adam optimizer 10-3. For multi-tasking experiments, we design hierarchical attention network use bidirectional GRUs consisting of 50 units, and tanh activation function. We use dropout layer of rate 0.5 and alternate training of the two tasks with two different optimization learning rates in order to make them synchronized to each other.  
 
We use Keras version 2.0.2 for training FT-MLP, MF-CNN and bi-GRU-Att and Tensorflow version 1.4.0 to implement multi-tasking models as they require lower-level handling of the loss function.

\section{Fine Grained results}
\label{app-5}
\subsection{Cross-lingual Document Classification}
\label{app-5-1}
Table \ref{cldc-all-tab} shows the fine grained analysis of the performance of different embeddings used for different neural architectures for document classification.

\subsection{Cross-lingual Churn Detection}
\label{app-5-2}
Table \ref{churn-all-tab} shows the fine grained analysis of the performance of different embeddings used for different neural architectures for churn detection.

\end{document}